\documentclass[10pt,twocolumn,letterpaper]{article}

\usepackage{cvpr}              

\usepackage[dvipsnames]{xcolor}

\usepackage{multirow}
\usepackage{float}
\usepackage{subcaption}
\usepackage{amsthm,amsmath,amssymb}
\usepackage{mathrsfs}

\definecolor{cvprblue}{rgb}{0.21,0.49,0.74}
\usepackage[pagebackref,breaklinks,colorlinks,citecolor=cvprblue]{hyperref}

\def\paperID{8357} 
\def\confName{CVPR}
\def\confYear{2024}

\title{PracticalDG: Perturbation Distillation on Vision-Language Models for Hybrid Domain Generalization}

\author{Zining Chen$^{1}$, Weiqiu Wang$^{1}$, Zhicheng Zhao$^{1,2,3}$\footnotemark[1]\, Fei Su$^{1,2,3}$, Aidong Men$^{1}$, Hongying Meng$^{4}$\\
$^{1}$The school of Artificial Intelligence, Beijing University of Posts and Telecommunications \\
$^{2}$Beijing Key Laboratory of Network System and Network Culture, China\\
$^{3}$Key Laboratory of Interactive Technology and Experience System Ministry \\of Culture and Tourism, Beijing, China\\
$^{4}$Brunel University Uxbridge \\
{\tt\small chenzn@bupt.edu.cn, $\{$wangweiqiu, zhaozc, sufei, menad$\}$@bupt.edu.cn, hongying.meng@brunel.ac.uk}
}

\begin{document}
\maketitle
\renewcommand{\thefootnote}{\fnsymbol{footnote}}
\footnotetext[1]{Corresponding author}
\footnotetext[2]{Source code is available at https://github.com/znchen666/HDG.}
\begin{abstract}
\vspace{-0.2cm}
Domain Generalization (DG) aims to resolve distribution shifts between source and target domains, and current DG methods are default to the setting that data from source and target domains share identical categories. Nevertheless, there exists unseen classes from target domains in practical scenarios. To address this issue, Open Set Domain Generalization (OSDG) has emerged and several methods have been exclusively proposed. However, most existing methods adopt complex architectures with slight improvement compared with DG methods. Recently, vision-language models (VLMs) have been introduced in DG following the fine-tuning paradigm, but consume huge training overhead with large vision models. Therefore, in this paper, we innovate to transfer knowledge from VLMs to lightweight vision models and improve the robustness by introducing Perturbation Distillation (PD) from three perspectives, including Score, Class and Instance (SCI), named SCI-PD. Moreover, previous methods are oriented by the benchmarks with identical and fixed splits, ignoring the divergence between source domains. These methods are revealed to suffer from sharp performance decay with our proposed new benchmark Hybrid Domain Generalization (HDG) and a novel metric $H^{2}$-CV, which construct various splits to comprehensively assess the robustness of algorithms. Extensive experiments demonstrate that our method outperforms state-of-the-art algorithms on multiple datasets, especially improving the robustness when confronting data scarcity.

\end{abstract}    
\vspace{-0.7cm}
\section{Introduction}
\begin{figure}
\centering
  \setlength{\belowcaptionskip}{-12pt}
  \setlength{\abovecaptionskip}{5pt}
  \includegraphics[width=0.97\linewidth]{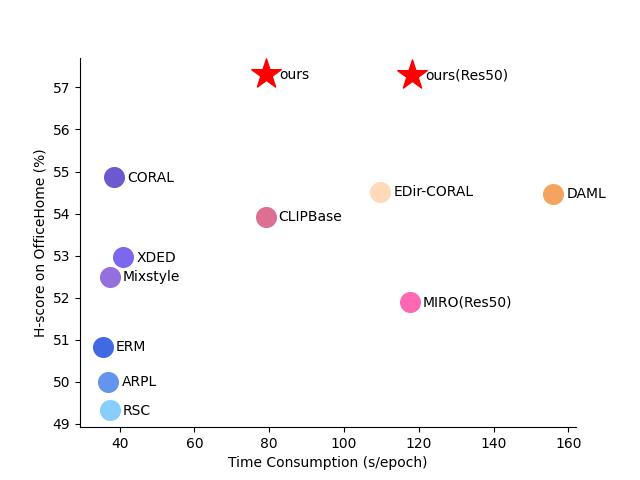}
    \caption{The balance between model performance and training time consumption. Model performance is evaluated on the average H-score of different splits based on the proposed HDG benchmark. Our method achieves superior performance with less training time compared with state-of-the-art (SOTA) methods in OSDG.  }
  \label{fig:time}
\end{figure}
\label{sec:intro}
\quad Deep learning has attained remarkable success on various downstream tasks in computer vision, typically under the presumption that both training and test samples are Independent and Identically Distributed (IID) with the same label space. However, real-world data often exhibits unpredictable distributions, leading to the failure of deep neural networks. To address such distribution shifts, Domain Generalization (DG) is first introduced to leverage data from multiple source domains to achieve generalization on unseen target domains, from the perspective of domain-invariant learning \cite{sun2016deep,shankar2018generalizing, lv2022causality,qu2023modality,lee2023decompose,guo2023domaindrop}, data augmentation \cite{zhou2020learning,xu2021fourier,zhou2021domain,kang2022style,choi2023progressive}, and learning strategies \cite{li2018domain, huang2020self,cha2021swad,wang2023sharpness,zhang2023flatness,chen2023instance}. However, it has been observed that most existing domain generalization methods assume a closed-set distribution, where the label space remains identical across the source and target domain. To address this limitation, Open Set Domain Generalization (OSDG) has emerged to resolve unseen classes from target domains \cite{katsumata2021open,shu2021open,zhu2021crossmatch,noguchi2023simple,wang2023generalizable,chen2023activate}. Nevertheless, most of these methods entail considerable computational costs but with little improvement that are impractical for real-world applications.


Recently, Vision-Language Models (VLMs) have shown powerful zero-shot transfer ability \cite{jia2021scaling,lei2021less,radford2021learning} on various downstream tasks. Then several researches have explored plausible solutions for VLMs on Out-of-Distribution (OOD) generalization \cite{cha2022domain,shu2023clipood,cho2023promptstyler,huang2023sentence,li2023distilling}. However, most solutions focus on fine-tuning or re-training the vision models to achieve high performance on the exclusive task, but inevitably suffer from large memory usage and computational costs. 
In contrast, our proposed perturbation distillation method can distill knowledge from large-scale VLMs to any lightweight vision models that introduces perturbation from three perspectives, including score, class and instance, named SCI-PD. 
As presented in Fig. \ref{fig:time}, our approach surpasses conventional DG and OSDG methods with a large margin. Compared with VLM-based fine-tuning methods, our method achieves superior performance with similar training time.  

\begin{figure}
\centering
  \setlength{\belowcaptionskip}{-5pt}
  \includegraphics[width=0.8\linewidth]{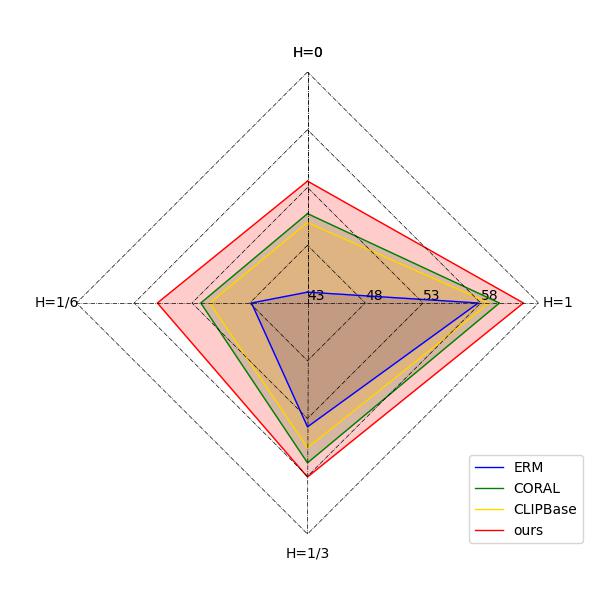}
    \caption{Illustration on the significance of the proposed HDG benchmark. Previous DG benchmarks are evaluated on a single split, producing unreliable conclusions for algorithms in practical usage. We claim that robust algorithms should possess stable performance on diverse data distributions.}
  \label{fig:HDG}
\end{figure}

Existing DG and OSDG methods are mostly evaluated on the benchmark that the label sets of multiple source domains are identical and fixed \cite{xu2021fourier,lv2022causality,shu2021open,wang2023generalizable}. Nevertheless, datasets derived from different resources in real-world applications merely contain a random subset of total classes, making it challenging to establish identical and fixed label sets across source domains. Therefore, to thoroughly evaluate the practical applicability of DG and OSDG methods, we propose a new benchmark called Hybrid Domain Generalization (HDG). As shown in Fig. \ref{fig:HDG}, HDG comprises of various splits to illustrate the diverse class discrepancy between source domains, producing reliable conclusions for algorithms in practical usage. Moreover, a novel metric ${\rm H^{2}}$-CV is proposed to measure the comprehensive robustness of the algorithms. 

In summary, this paper aims to enhance the practicality of domain generalization from the perspective of algorithm, benchmark and metric, which can be summarized as follows,
\begin{itemize}
    \item We propose a more practical method, called SCI-PD based on VLMs to address the OSDG task. We dismiss the fine-tuning or re-training paradigm, and design perturbation from score, class and instance to distill lightweight vision models. To the best of our knowledge, we are the first to transfer knowledge from VLMs to lightweight vision models for OSDG.
    \item We propose a more practical task of domain generalization, called Hybrid Domain Generalization (HDG), which is open set and the label sets of different source domains are disparate and diverse. Meanwhile, a new evaluation metric ${\rm H^{2}}$-CV is proposed to comprehensively assess model robustness. 
    \item Experimental results on different HDG benchmarks manifest the superior performance of our method in comparison with previous DG, OSDG and VLM-based methods. SCI-PD not only achieves state-of-the-art performance on accuracy of source and target classes, but also shows powerful robustness under the proposed metric ${\rm H^{2}}$-CV. 
\end{itemize}

\section{Related Work}
\quad \textbf{Domain Generalization} intends to train a model from multiple source domains and migrate to arbitrary unseen target domains. Currently, DG methods can be roughly divided into three categories, including domain-invariant learning \cite{sun2016deep,shankar2018generalizing, lv2022causality,qu2023modality,lee2023decompose,guo2023domaindrop}, data augmentation \cite{zhou2020learning,xu2021fourier,zhou2021domain,kang2022style,choi2023progressive}, and learning strategies \cite{li2018domain, huang2020self,cha2021swad,wang2023sharpness,zhang2023flatness,chen2023instance}. Most methods have achieved outstanding performance, but inevitably indulge complex architectures and require extensive training strategies that are impractical in real-world scenarios. In this context, appropriate perturbation on instance and feature has succeeded with little extra computational costs \cite{xu2021fourier,zhou2021domain}. From this perspective, we propose a novel perturbation distillation method on vision-language models that can be transferred to any lightweight vision models.

\textbf{Open Set Domain Generalization} has recently been proposed as a promising solution to tackle the impracticality of closed-set distribution in domain generalization. To the best of our knowledge, there are only few related works that specifically address this issue \cite{katsumata2021open,shu2021open,zhu2021crossmatch,noguchi2023simple,wang2023generalizable,chen2023activate}. \cite{shu2021open} pioneers the formation of diverse label sets across source and target domains, introducing a feature-level augmentation and a label-level distillation with meta-learning. \cite{katsumata2021open} designs a decoupling loss to refine the feature representation of unknown samples, thereby constructing a distinguishable feature space. Recent study from \cite{noguchi2023simple} acknowledges the computational costs of \cite{shu2021open} and proposes to integrate its techniques with conventional DG methods. \cite{chen2023activate} designs a post hoc modification on test-time unknown rejection to discriminate test data for safe deployment.  \cite{wang2023generalizable} considers gradient matching across both inter-class and inter-domain splits via meta-learning, yet it requires an identical label distribution for source domains. Furthermore, \cite{zhu2021crossmatch} proposes a more challenging scenarios, Open-Set Single Domain Generalization (OS-SDG), which the model is exclusively trained on a single source domain. It leverages adversarial learning to simulate the data distribution of unknown classes. To sum up, while the aforementioned methods have shed light on OSDG, they are concurrently constrained on model robustness and computational costs.


\textbf{Vision-Language Models} have achieved great advancements in pretraining on large-scale image-text datasets \cite{jia2021scaling,lei2021less,radford2021learning}. Recently, the Contrastive Language-Vision Pre-training \cite{radford2021learning} method achieves remarkable performance on downstream tasks. Most recent studies \cite{zhou2022learning,wortsman2022robust,gao2023clip} adheres to the fine-tuning and re-training paradigm tailored to specific downstream tasks. Inspired by these studies, several methods \cite{cha2022domain,shu2023clipood,cho2023promptstyler,huang2023sentence} have concentrated on how to transfer knowledge from CLIP models to OOD scenarios. \cite{cha2022domain} leverages mutual information from vision-language models to guide the training of the task-specific model. \cite{shu2023clipood} designs a semantic training objective with a novel optimization strategy from the perspective of fine-tuning CLIP models. \cite{cho2023promptstyler} proposes diverse learnable vectors as pseudo-words to synthesize novel styles in prompts for source-free domain generalization. \cite{huang2023sentence} considers to enforce the image embedding from smaller models closer to the corresponding text embedding from VLMs for DG task. Nevertheless, there are no explorations on distillation of CLIP models to lightweight vision models for OSDG.

\textbf{Knowledge Distillation} (KD) has been studied in early stage to transfer knowledge from teacher models to student models \cite{hinton2015distilling,park2019relational}. 
Techniques in KD have evolved into various aspects, such as self-distillation that has achieved comparable performance \cite{zhang2019your,yun2020regularizing}. Recently, distillation on CLIP models have become a promising solution on diverse downstream tasks. \cite{gu2021open,yang2022cross,zheng2023preventing,shu2023clipood} concentrate on fine-tuning vision models for uni-modal tasks, while \cite{fang2021clip2video,dai2022enabling,pei2023clipping,sun2023dime} focuse on multi-modal tasks. Despite the aforementioned success of KD, there are only few techniques that specifically address domain generalization. \cite{wang2021embracing} proposes a teacher-to-student distillation network with a regularization term on gradients. \cite{lee2022cross} introduces a novel training objective that imposes penalties on discrepancies between single logits and ensembled counterparts. \cite{sultana2022self} proposes self-distillation on classification token between random intermediate transformer blocks and final blocks, but is exclusively designed for Vision Transformers (ViT) \cite{dosovitskiy2020image} that limits practicality.

\section{Method}
\label{sec:method}
\begin{figure*}
\centering
  \setlength{\belowcaptionskip}{-3pt}
  \includegraphics[width=0.86\linewidth,height=0.33\textheight]{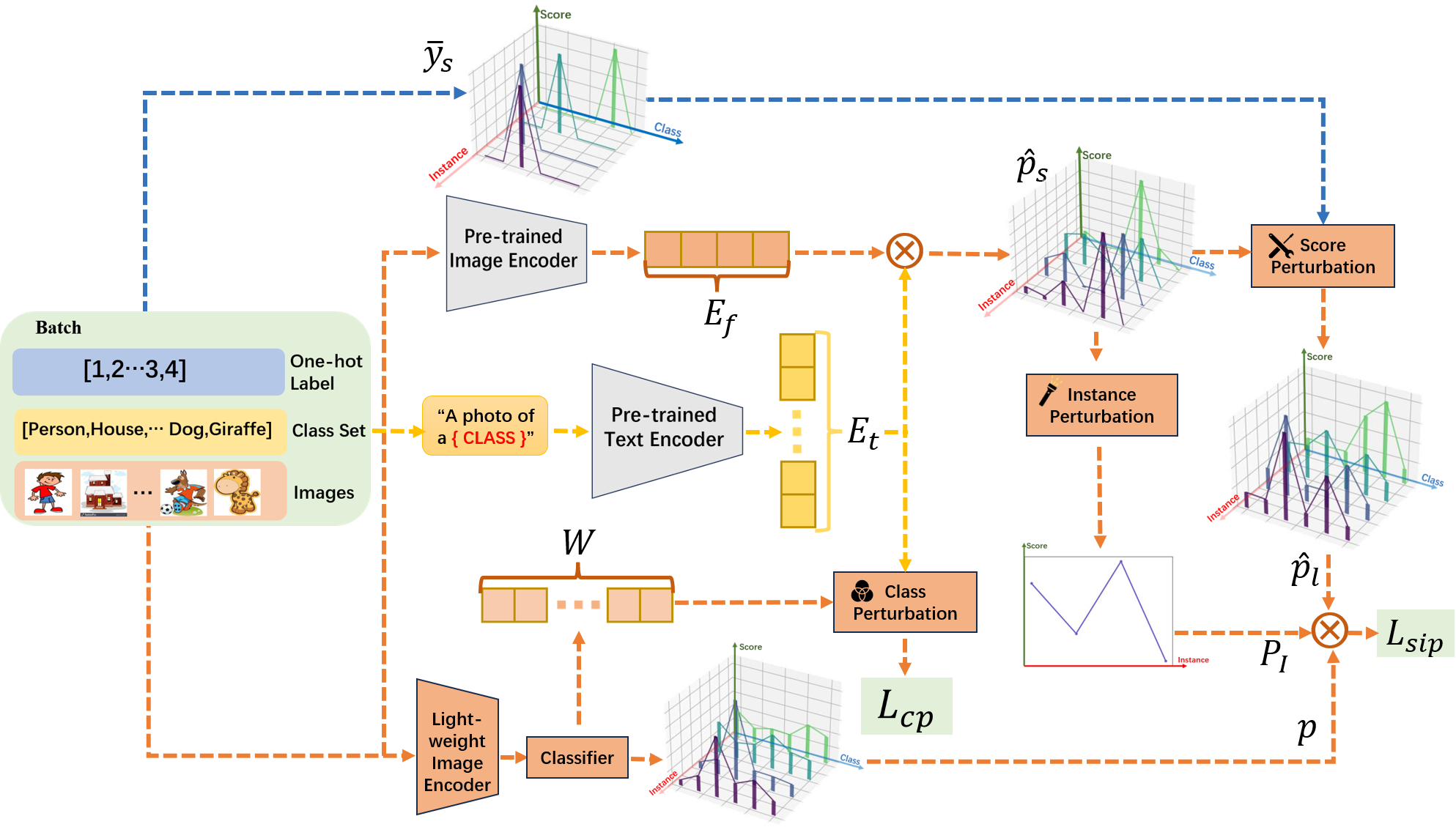}
    \caption{The overall framework of our method SCI-PD, including Score Perturbation (SP), Instance Perturbation (IP) and Class Perturbation (CP). SP saturates GT information into the similarity scores from CLIP to exploit semantics. IP excavates underlying semantics in instances via the weight distribution. CP saturates semantics from pretrained text embeddings to the class weights of the classifier. }
  \label{fig:Framework}
\end{figure*}

\quad In this section, we first introduce the preliminaries of OSDG and CLIP model. Then a detailed description on our method SCI-PD is presented, as shown in Fig. \ref{fig:Framework}. 
\subsection{Preliminaries}
\quad \textbf{Open Set Domain Generalization.} Suppose there are multiple source domains $\mathcal{D}_{1}, \mathcal{D}_{2}, ..., \mathcal{D}_{S}$ for training, where each domain $\mathcal{D}_{s}=\left\{(x_{s}, y_{s})\right\}_{s=1}^{n_{s}}$ consists of  $n_{s}$ data-label pairs with unique label set $\mathcal{C}_{s}$. Also, there are certain target domains where domain $\mathcal{D}_{t}=\left\{(x_{t}, y_{t})\right\}_{t=1}^{n_{t}}$ has diverse label set $\mathcal{C}_{t}$. Assume the union of label sets from source domains $\mathcal{C}_{1}, \mathcal{C}_{2}, ..., \mathcal{C}_{S}$ as $\mathcal{C}$, then $\mathcal{C}_{a}=\mathcal{C}  \cup  \mathcal{C}_{t}$ represents the total label sets, whereas $\mathcal{C}_{u}=\mathcal{C}_{a} \setminus \mathcal{C}$ is label sets of unknown classes. The goal of OSDG is to train models on source domains, and generalize well on the unseen target domains with unknown classes, where the target sample $x_{t}$ should be classified as the correct class if it belongs to $\mathcal{C}$ or should be labeled as ``unknown" if it belongs to $\mathcal{C}_{u}$. Similar to domain generalization, no data and label from target domains are available for training.

\textbf{CLIP model \cite{radford2021learning}.} The CLIP model consists of an image encoder $f_{I}$ and a text encoder $f_{T}$. Previous studies on CLIP model for zero-shot inference on downstream tasks usually adopt the following procedure. First, each target class $c \in \mathcal{C}_{t}$ is transformed using a template such as ``a photo of a $\left\{c\right\}$". Then, the text encoder transforms the class tokens from all classes into the text embeddings $E_{t}=[e_{ti}]_{i=1}^{N}$, while an image encoder simultaneously encodes the input images into the image embeddings $E_{f}=[e_{fi}]_{i=1}^{B}$, where $N$ is the number of known classes and $B$ denotes the number of instances in a batch. Finally, the cosine similarity of the image embedding and the text embedding is calculated as $s_{i}=\langle{e_{ti}, e_{fi}}\rangle$, and the class with the maximum cosine similarity is the predicted label of the image.

As recent studies focus on how to finetune CLIP for downstream tasks, we observe that a simple variant of cross-entropy loss is employed with remarkable achievement \cite{hinton2015distilling}. Specifically, the similarity between each image embedding with different text embeddings is calculated as $s=\left\{s_{1},s_{2},...s_{N}\right\}$, and is then normalized using Softmax function,
\vspace{-0.3cm}
\begin{equation}
\small
    \hat{p}_{s} = \mathscr{S}(s;\lambda) = \frac{exp(s_{i}/\lambda)}{\sum_{i=1}^{N}exp(s_{i}/\lambda)}
\label{softmax}
\end{equation}
where $\mathscr{S}$ is Softmax function and $\lambda$ is the conventional temperature in CLIP. Note that $\lambda$ equals to 1 when omitted. Then, the normalized similarity is leveraged to guide the classification process, 
\begin{equation}
\small
    \mathcal{L}_{base} = CE(p,\hat{p}_{s})
\label{base}
\end{equation}
where $p$ represents the output from the downstream image encoder and the classifier, and $CE(\cdot;\cdot)$ represents the cross-entropy loss. Note that we denote this method as $\rm CLIPBase$ for abbreviation. 

\subsection{Score Perturbation}
\quad The success of $\rm CLIPBase$ makes us reconsider the difference between $\rm CLIPBase$ and the baseline method ERM \cite{koltchinskii2011oracle} in DG. 
CLIPBase takes the similarity $\hat{p}_{s}$ from CLIP as supervision. We observe that the distribution of $\hat{p}_{s}$ contains affluent semantics which are essential for domain-invariant learning, but it inevitably introduces redundant noise from incorrect predictions. ERM merely utilizes the Ground-Truth (GT) label for supervision, but rigidly restricts semantics and introduces domain-specific information that mislead model convergence. 
Consequently, we introduce Score Perturbation (SP) to balance semantics from CLIP and GT labels. 

Specifically, suppose a sample from source domains as $x_{s}$ whose GT label is $y_{s}$ and the similarity from CLIP is $\hat{p}_{s}\in\mathbb{R}^{1 \times N}$. We obtain the index $y_{c}$ of the maximum similarity $\hat{p}_{s}$ as the predicted label from CLIP. Obviously, there appears to be misclassified samples that $y_{c} \neq y_{s}$. Thus, we design masks that follows the formula:
\begin{equation}
\small
    mask = \left\{
    \begin{aligned}
    & \mathbf{1}, \quad & y_{c} \neq y_{s} \\
    & \mathbf{0}, \quad & y_{c} = y_{s} \\
    \end{aligned}
    \quad \in \mathbb{R}^{1 \times N}
    \right.
\end{equation}
where $\mathbf{1}$ and $\mathbf{0}$ are all-ones matrix and all-zeros matrix, respectively. Then we saturate the GT labels into the similarity from CLIP based on the mask. Suppose the maximum similarity as $\hat{p}_{s,max}$ and the one-hot label of $y_{s}$ as $\overline{y}_{s} \in \mathbb{R}^{1 \times N}$. We establish the score perturbation $P_{L}$ as,
\begin{equation}
\small
    P_{L} = mask \odot (\hat{p}_{s,max} \times \overline{y}_{s}) \in\mathbb{R}^{1 \times N}
\end{equation}
where $\times$ denotes the cross product and $\odot$ is the Hadamard product. Then we add this perturbation to the similarity $\hat{p}_{s}$ with a $\tau$-Softmax for normalization,
\begin{equation}
\small
    \hat{p}_{l} = \mathscr{S}(\hat{p}_{s} + P_{L};\tau)
\label{lpsoftmax}
\end{equation}
The proposed SP has two-fold advantages. Firstly, SP remains the distribution of $\hat{p}_{s}$ that successfully preserves the semantics that boost the domain-invariant learning. Secondly, SP saturates perturbation from accurate GT labels that suppress semantic noises from CLIP.

\subsection{Instance Perturbation}
\quad The similarity from CLIP is to quantify the relations between an image and all the classes. Consequently, instances with sharp distribution of $\hat{p}_{s}$  manifest low similarity relative to other classes, whose semantics are scarce for domain-invariant learning. In contrast, a more uniform distribution of $\hat{p}_{s}$ suggests that the instance share more commonalities with other classes, in which the abundant semantics are implicitly included. 
From this perspective, we observe that the original objective $\mathcal{L}_{base}$ in Eq.~\ref{base} equally address all instances that semantics from a more uniform distribution are constrained. 

We propose instance perturbation to excavate more underlying semantics. The maximum similarity $\hat{p}_{s,max}$ represents the certainty of CLIP model on the classification of the image $x_{s}$, and lower certainty stands for a higher possibility for affluent semantics, and vice versa. Consequently, we use a exponential reciprocal as the perturbation, 
\begin{equation}
\small
    P_{I} = (\frac{1}{\hat{p}_{s,max}})^{\alpha}
\label{ip}
\end{equation}
Then, the total classification loss can be defined with the saturation on score perturbation and instance perturbation as follows,
\begin{equation}
\small
    \mathcal{L}_{sip} = P_{I} \cdot CE(p,\hat{p}_{l})
\label{cefinal}
\end{equation}

\subsection{Class Perturbation}
\quad CLIP is trained with the objective of a contrastive loss between the image and the text modality. However, most downstream vision tasks merely use the pretrained image encoder from CLIP with a new classifier as the network \cite{wortsman2022robust,cha2022domain}. Thus, when fine-tuning the downstream vision model, the alignment between two modalities are broken that deteriorates the performance \cite{shu2023clipood}. From this perspective, we design class perturbation to saturate the semantics from pretrained text encoder to the classifier.

Let $W\in\mathbb{R}^{N\times C_{w}}$ denotes the weight of the classifier, and first the L2-norm is applied on $E_{t}$ and $W$. 
Then, we design the class perturbation as the similarity of $E_{t}$,
\begin{equation}
\small
     P_{C} = E_{t} \cdot E_{t}^{T}
\end{equation}
Next, we add the perturbation by pulling the similarity between $E_{t}$ and $W$ closer to the perturbation $P_{C}$ with loss $\mathcal{L}_{cp}$, 
\begin{equation}
\small
    S_{t,w} = W \cdot E_{t}^{T}
\end{equation}
Finally, the class perturbation loss is:
\begin{equation}
\small
\begin{aligned}
        L_{cp} &= CE(\mathscr{S}(S_{t,w}), \mathscr{S}(P_{C}))\\
    & + CE(\mathscr{S}(S_{t,w}^T), \mathscr{S}(P_{C}^T))
\end{aligned}
\end{equation}

\subsection{Train and Inference}
\quad Combining the two losses, the final training objective is:
\begin{equation}
\small
     L_{SCI-PD} = L_{sip} + \beta \times L_{cp}
\end{equation}
where $\beta$ is the trade-off hyper-parameter between the classification loss and the class perturbation loss. 

For inference, all algorithms follow the same procedure proposed in \cite{shu2021open,noguchi2023simple,wang2023generalizable}.

\section{Hybrid Domain Generalization}

\quad \textbf{Hybridness.} We start from the definition of hybridness $\mathcal{H}$ that illustrates the insight of the proposed HDG. As source domains derived from diverse resources are difficult to maintain identical label sets, $\mathcal{H}$ is designed to measure the discrepancy between label sets of source domains. Specifically, let the intersection of the label sets from two source domains as $\mathcal{C}_{i,j} = \mathcal{C}_{i} \cap \mathcal{C}_{j}$, and all combinations of two source domains as $U$ whose total number of combination pairs is,
\begin{equation}
\small
    |U| = C_M^2 = \frac{M(M-1)}{2}
\end{equation}
where $|\cdot|$ denotes the number of elements and $M$ is the number of source domains. Thus, the hybridness $\mathcal{H}$ is defined as,
\begin{equation}
\small
    \mathcal{H} = \frac{\sum_{(i,j)}^U |\mathcal{C}_{i,j}|}{N|U|}
\end{equation}
The hybridness can simultaneously reflect the overlap between source domains and the severity of data scarcity. A smaller $\mathcal{H}$ signifies a less overlap between source domains and a greater data scarcity. 

\textbf{Conventional Benchmarks.} With the definition of hybridness, all the conventional benchmarks are unified and evaluated only under a single scenario. Specifically, the OSDG benchmark was initially introduced in \cite{shu2021open} that aims to evaluate the accuracy across both known and unknown categories. However, $\mathcal{H}$ is fixed that is not sufficient to evaluate model robustness. The most recent work \cite{wang2023generalizable} on OSDG is constrained to identical label space across source domains, which is less persuasive because it mandates $\mathcal{H}=1$. Furthermore, \cite{zhu2021crossmatch} proposes an OS-SDG benchmark, which is less challenging compared with the situation when $\mathcal{H}=0$. Besides, it has no appropriate metric to evaluate model robustness. 

\textbf{HDG Benchmark.}
In practice, label sets of different source domains are disparate and diverse. 
However, previous methods are oriented by conventional benchmarks with a fixed hybridness that are not practical, and exhibit significant performance degradation when hybridness changes or even are infeasible for implementation under other hybridness.
Therefore, we modify the hybridness to build the HDG benchmark. Specifically, we pre-set four different representative $\mathcal{H}$ to establish four splits, including $0, \frac{1}{2M}, \frac{1}{M}, 1$ (detailed splits for each dataset are presented in the supplementary material). Meanwhile, we propose a new evaluation metric ${\rm H^{2}}$-CV to comprehensively assess the robustness of the algorithms based on the H-score from different $\mathcal{H}$. Specifically, ${\rm H^{2}}$-CV utilizes the coefficient of variation in Statistics to evaluate the dispersion of a distribution. Suppose the set of discrete values of H-score on different $\mathcal{H}$ is $S$, then the formula of ${\rm H^{2}}$-CV is,
\begin{equation}
\small
    {\rm H^{2}}\text{-CV} = \frac{\sigma(S)}{\overline{S}} \times 100\%
\end{equation}
where $\sigma(S)$ is the standard deviation value and $\overline{S}$ is the mean value. Consequently, an algorithm with low ${\rm H^{2}}$-CV means a small $\sigma(S)$ and a large $\overline{S}$ that is considered as robust. Meanwhile, we adopt two other evaluation metrics: Top-1 accuracy and H-score \cite{bucci2020effectiveness}, which have been widely utilized to assess the accuracy on known categories and unknown categories. 

\section{Experiments}
\subsection{Experiment Setup}
\quad We conduct experiments on three datasets for domain generalization, including PACS \cite{li2017deeper}, OfficeHome \cite{venkateswara2017deep} and DomainNet \cite{peng2019moment} on the proposed HDG benchmark. 
We adopt CLIP model \cite{radford2021learning} with ViT-B/16 \cite{dosovitskiy2020image} as the image encoder, and select ResNet18 \cite{he2016deep} pretrained on ImageNet \cite{deng2009imagenet} as the lightweight vision model if not specified. We follow the leave-one-domain-out evaluation protocol on all datasets, and select the model with the best accuracy on validation splits for testing. The evaluation metrics are Top-1 accuracy, H-score and the proposed ${\rm H^{2}}$-CV. More dataset and implementation details can be found in the supplementary material. 
\subsection{Comparison with State-of-the-Art Methods}
\quad We compare the proposed method with 12 methods, including ERM \cite{koltchinskii2011oracle}, CORAL \cite{sun2016deep}, MMD \cite{li2018domain}, RSC \cite{huang2020self}, MixStyle \cite{zhou2021domain}, CIRL \cite{lv2022causality}, XDED \cite{lee2022cross}, RISE \cite{huang2023sentence} for closed-set DG; ARPL \cite{chen2021adversarial}, DAML \cite{shu2021open}, EDir-CORAL \cite{noguchi2023simple}, MEDIC \cite{wang2023generalizable} for OSDG. For a fair comparison, we conduct the baseline of VLM-based method CLIPBase as clarified in Section \ref{sec:method}. Note that the accuracy and H-score on different $\mathcal{H}$ is the average of all domains following the leave-one-domain-out protocol. We report the detailed results on each domain in the supplementary material. 

\begin{figure}[!ht]
  \centering
  \subcaptionbox{Art}
  {\includegraphics[width=0.49\linewidth]{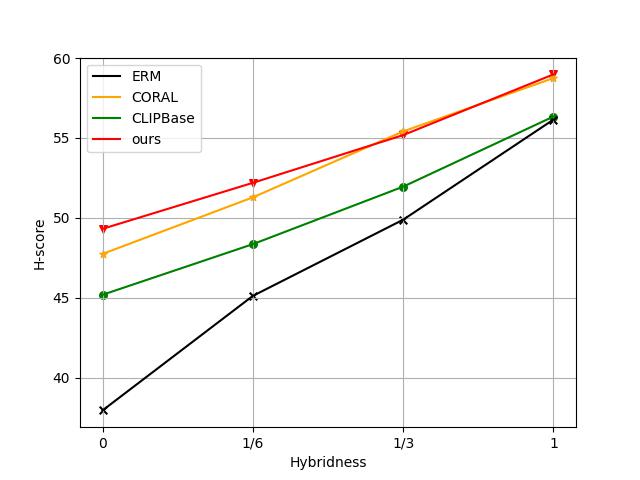}}
  \subcaptionbox{Clipart}{\includegraphics[width=0.49\linewidth]{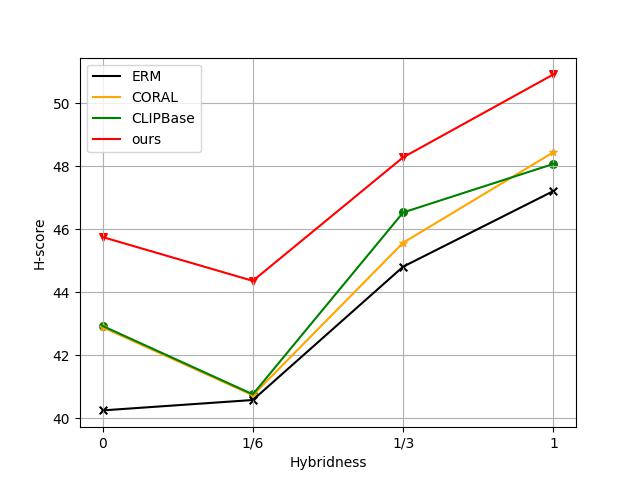}}
  \subcaptionbox{Product}{\includegraphics[width=0.49\linewidth]{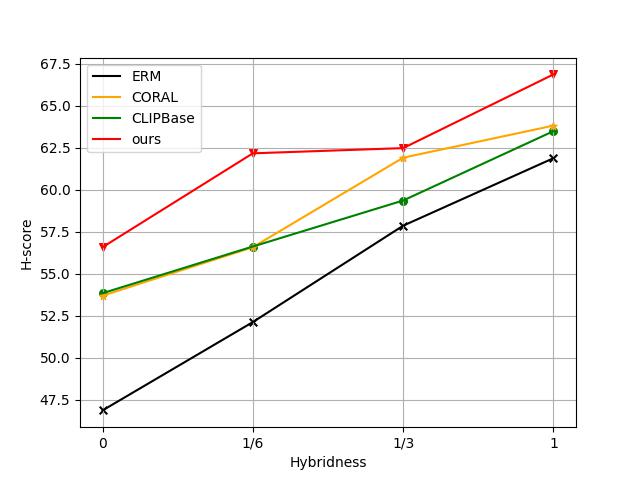}}
  \subcaptionbox{Real World}{\includegraphics[width=0.49\linewidth]{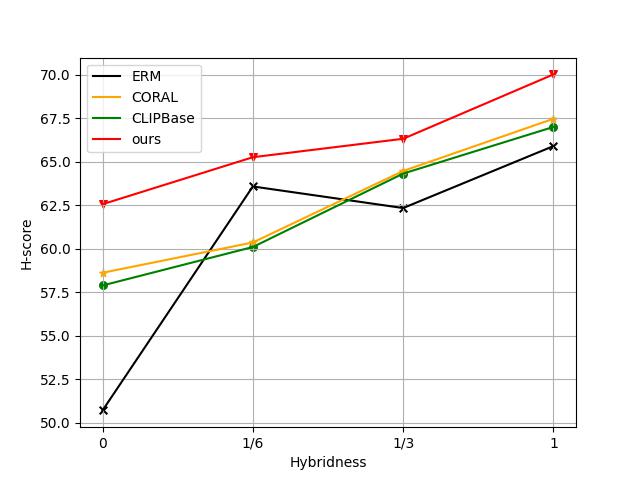}}
  \caption{H-score on different domains under diverse hybridness $\mathcal{H}$ for OfficeHome.}
  \label{fig:domain}
\end{figure}

\textbf{OfficeHome \cite{venkateswara2017deep}.}
As shown in Table \ref{tab:office}, it can be observed that our method SCI-PD surpasses all other SOTA methods on the three metrics. Concretely, our method exceeds the SOTA method XDED \cite{lee2022cross} with 3.91$\%$ on accuracy and 4.37$\%$ on H-score. Meanwhile, the metric ${\rm H^{2}}$-CV is capable to show the robustness of the algorithms. XDED can achieve comparable performance when $\mathcal{H}=1$, but performance degrades on other settings that results in a high ${\rm H^{2}}$-CV with poor robustness. CORAL \cite{sun2016deep} is considered as a robust algorithm with a low ${\rm H^{2}}$-CV of 6.47$\%$, and the ${\rm H^{2}}$-CV of EDir-CORAL \cite{noguchi2023simple} decreases compared with DAML \cite{shu2021open}. 
Compared with the baseline method CLIPBase on VLMs, SCI-PD can improve 3.62$\%$ and 3.41$\%$ on accuracy and H-score and a further 1.20$\%$ improvement on ${\rm H^{2}}$-CV that proves the effectiveness of the proposed method. Moreover, although RISE is a recent CLIP-based DG algorithm, it performs 7.43$\%$ and 5.92$\%$ worse than SCI-PD where the performance drops significantly when $\mathcal{H}=0$. Furthermore, Fig. \ref{fig:domain} illustrates that the proposed method holds the capability for enhancing performance on all domains rather than a single domain. Especially the variance in the Art domain evidently validates the robustness of our method that the discrepancy between different $\mathcal{H}$ is small.

\begin{table*}[!ht]
\small
\setlength{\abovecaptionskip}{-5pt} 
\setlength{\belowcaptionskip}{-3pt} 
\caption{Comparison of state-of-the-art methods on Acc ($\%$), H-score ($\%$) and ${\rm H^{2}}$-CV ($\%$) for OfficeHome.}
\label{tab:office}
\begin{center}
\setlength{\tabcolsep}{5pt}
\begin{tabular}{c|cc|cc|cc|cc|cc|c}
\hline
\multirow{2}{*}{Method} & \multicolumn{2}{c|}{$\rm \mathcal{H}=0$} & \multicolumn{2}{c|}{$\rm \mathcal{H}=1/6$} & \multicolumn{2}{c|}{$\rm \mathcal{H}=1/3$} & \multicolumn{2}{c|}{$\rm \mathcal{H}=1$} & \multicolumn{2}{c|}{Average} & \multirow{2}{*}{${\rm H^{2}}$-CV ($\downarrow$)}\\
\cline{2-11}
& Acc & H-score & Acc & H-score & Acc & H-score & Acc & H-score & Acc & H-score\\
\hline
ERM \cite{koltchinskii2011oracle} & 46.09 & 43.94 & 50.64 & 47.85 & 59.92 & 53.73 & 65.55 & 57.79 & 55.55 & 50.83 & 10.47\\
ARPL \cite{chen2021adversarial} & 44.31 & 43.28 & 48.62 & 46.68 & 57.68 & 53.06 & 63.58 & 56.95 & 53.55 & 49.99 & 10.66\\
RSC \cite{huang2020self} & 41.95 & 41.59 & 47.21 & 45.98 & 56.94 & 52.47 & 63.59 & 57.25 & 52.42 & 49.32 & 12.15\\
MMD \cite{li2018domain} & 50.77 & 47.30 & 53.62 & 49.37 & 61.21 & 54.69 & 65.53 & 58.32 & 57.78 & 52.42 & 8.28\\
Mixstyle \cite{zhou2021domain} & 48.75 & 46.30 & 53.11 & 49.97 & 61.15 & 54.68 & 66.89 & 59.00 & 57.47 & 52.49 & 9.13\\
CORAL \cite{sun2016deep} & 55.14 & 50.74 & 57.22 & 52.24 & 63.37 & 56.84 & 67.68 & 59.62 & 60.85 & 54.86 & 6.47\\
DAML \cite{shu2021open} & 51.64 & 48.60 & 54.95 & 51.80 & 62.22 & 56.86 & 67.36 & 60.61 & 59.04 & 54.46 & 8.46\\
EDir-CORAL \cite{noguchi2023simple} & 52.01 & 49.07 & 55.09 & 51.93 & 61.90 & 56.76 & 66.81 & 59.89 & 58.95 & 54.42 & 7.70\\
XDED \cite{lee2022cross} & 49.11 & 46.25 & 53.32 & 50.26 & 61.96 & 55.65 & 66.89 & 59.70 & 57.82 & 52.96 & 9.67\\
RISE \cite{huang2023sentence} & 43.58 & 43.40 & 48.59 & 48.08 & 59.20 & 54.69 & 65.82 & 59.46 & 54.30 & 51.41 & 11.94\\
\hline
CLIPBase & 52.38 & 49.96 & 54.48 & 51.46 & 60.80 & 55.54 & 64.78 & 58.73 & 58.11 & 53.92 & 6.40\\
\textbf{SCI-PD} & \textbf{56.94} & \textbf{53.55} & \textbf{58.25} & \textbf{56.00} & \textbf{63.66} & \textbf{58.07} & \textbf{68.08} & \textbf{61.70} & \textbf{61.73} & \textbf{57.33} & \textbf{5.20}\\
\hline
\end{tabular}
\end{center}
\end{table*}

\textbf{PACS \cite{li2017deeper}.}
As PACS is a relatively simple benchmark with merely 7 classes, we observe that the SOTA methods, such as XDED \cite{lee2022cross} and CIRL \cite{lv2022causality} on closed-set DG, achieve comparable performance with VLM-based method under the setting $\mathcal{H}=1$ in Table \ref{tab:pacs}. However, they suffer from 18.76$\%$ and 14.49$\%$ decay on H-score compared with SCI-PD when $\mathcal{H}=1/6$. Meanwhile, results on ${\rm H^{2}}$-CV shows that the variation of the robustness on different methods is high. As the SOTA methods CIRL and XDED achieve comparable performance on accuracy and H-score, CIRL obviously has better robustness. Moreover, RISE \cite{huang2023sentence} merely performs well when $\mathcal{H}=1$, but suffers from 6.29$\%$ on ${\rm H^{2}}$-CV that demonstrates the low robustness. MEDIC \cite{wang2023generalizable} is even constrained with $\mathcal{H}=1$ that cannot be employed on other settings, limiting its practicality in real-world applications. From this perspective, SCI-PD exceeds all methods on the robustness and stability. 
\begin{table*}[!ht]
\small
\setlength{\abovecaptionskip}{-5pt} 
\setlength{\belowcaptionskip}{-3pt} 
\caption{Comparison of state-of-the-art methods on Acc ($\%$), H-score ($\%$) and ${\rm H^{2}}$-CV ($\%$) for PACS.}
\label{tab:pacs}
\begin{center}
\setlength{\tabcolsep}{5pt}
\begin{tabular}{c|cc|cc|cc|cc|cc|c}
\hline
\multirow{2}{*}{Method} & \multicolumn{2}{c|}{$\rm \mathcal{H}=0$} & \multicolumn{2}{c|}{$\rm \mathcal{H}=1/6$} & \multicolumn{2}{c|}{$\rm \mathcal{H}=1/3$} & \multicolumn{2}{c|}{$\rm \mathcal{H}=1$} & \multicolumn{2}{c|}{Average} & \multirow{2}{*}{${\rm H^{2}}$-CV ($\downarrow$)}\\
\cline{2-11}
& Acc & H-score & Acc & H-score & Acc & H-score & Acc & H-score & Acc & H-score\\
\hline
ERM \cite{koltchinskii2011oracle} & 35.85 & 28.97 & 52.30 & 45.57 & 71.38 & 61.20 & 82.81 & 70.52 & 60.59 & 51.56 & 30.64\\
ARPL \cite{chen2021adversarial} & 37.22 & 30.37 & 54.50 & 46.96 & 72.03 & 61.59 & 81.91 & 68.24 & 61.41 & 51.79 & 28.13\\
Mixstyle \cite{zhou2021domain} & 43.14 & 28.00 & 60.76 & 48.71 & 76.95 & 64.88 & 84.11 & 71.22 & 66.24 & 53.20 & 31.41\\
MMD \cite{li2018domain} & 38.12 & 37.89 & 56.62 & 50.89 & 73.91 & 64.24 & 80.21 & 69.36 & 62.21 & 55.59 & 22.04\\
CORAL \cite{sun2016deep} & 39.85 & 37.59 & 60.49 & 48.27 & 72.68 & 62.96 & 82.27 & 69.02 & 63.82 & 54.46 & 22.62\\
EDir-CORAL \cite{noguchi2023simple} & 41.25 & 37.10 & \textbf{68.81} & 56.74 & 78.49 & 67.27 & 84.48 & 72.04 & 68.26 & 58.29 & 23.04\\
XDED \cite{lee2022cross} & 36.60 & 16.71 & 51.55 & 38.11 & 74.00 & 61.50 & 84.23 & 71.17 & 61.60 & 46.87 & 45.15\\
CIRL \cite{lv2022causality} & \textbf{50.32} & 35.27 & 61.46 & 42.38 & 72.33 & 45.07 & 85.29 & 62.72 & 67.35 & 46.36 & 21.79\\
RISE \cite{huang2023sentence} & 39.51 & 34.51 & 59.87 & 53.14 & 75.59 & 70.14 & 82.10 & \textbf{75.71} & 64.27 & 58.38 & 27.56\\
MEDIC \cite{wang2023generalizable} & - & - & - & - & - & - & \textbf{86.20} & 71.47 & - & - & -\\
\hline
CLIPBase & 43.91 & 38.23 & 64.32 & 55.14 & 79.16 & 69.20 & 84.22 & 72.89 & 67.90 & 58.87 & 23.15\\
\textbf{SCI-PD} & 48.69 & \textbf{41.93} & 66.58 & \textbf{56.87} & \textbf{80.13} & \textbf{71.03} & 85.25 & 75.03 & \textbf{70.16} & \textbf{61.21} & \textbf{21.27}\\
\hline
\end{tabular}
\end{center}
\end{table*}

\textbf{DomainNet \cite{peng2019moment}.}
We refer to the results in DomainBed \cite{gulrajani2020search} and discover that ERM \cite{koltchinskii2011oracle} is a strong baseline in DomainNet, surpassing most DG methods. Thus, we merely implement recent SOTA methods and results in Table \ref{tab:domainnet} show a slight improvement with a maximum of 0.48$\%$ on H-score. Nevertheless, CLIPBase exceeds ERM with 2.50$\%$ on H-score and 9.60$\%$ on ${\rm H^{2}}$-CV that proves the powerful zero-shot ability of VLMs. As DomainNet is a challenging dataset which is difficult to improve performance, SCI-PD shows a 2.73$\%$ and 2.14$\%$ improvement on accuracy and H-score compared with CLIPBase. Moreover, the performance of SCI-PD when $\mathcal{H}=1/10$, can surpass all conventional DG methods when $\mathcal{H}=1/5$, which is capable to resolve the issue of data scarcity. 
\begin{table*}[ht]
\small
\setlength{\abovecaptionskip}{-5pt} 
\setlength{\belowcaptionskip}{-3pt} 
\caption{Comparison of state-of-the-art methods on Acc ($\%$), H-score ($\%$) and ${\rm H^{2}}$-CV ($\%$) for DomainNet.}
\label{tab:domainnet}
\begin{center}
\setlength{\tabcolsep}{5pt}
\begin{tabular}{c|cc|cc|cc|cc|cc|c}
\hline
\multirow{2}{*}{Method} & \multicolumn{2}{c|}{$\rm \mathcal{H}=0$} & \multicolumn{2}{c|}{$\rm \mathcal{H}=1/10$} & \multicolumn{2}{c|}{$\rm \mathcal{H}=1/5$} & \multicolumn{2}{c|}{$\rm \mathcal{H}=1$} & \multicolumn{2}{c|}{Average} & \multirow{2}{*}{${\rm H^{2}}$-CV ($\downarrow$)}\\
\cline{2-11}
& Acc & H-score & Acc & H-score & Acc & H-score & Acc & H-score & Acc & H-score\\
\hline
ERM \cite{koltchinskii2011oracle} & 17.21 & 21.94 & 27.08 & 31.07 & 29.98 & 33.71 & 38.69 & 40.70 & 28.24 & 31.86 & 21.10\\
ARPL \cite{chen2021adversarial} & 17.10 & 21.78 & 27.23 & 30.97 & 30.46 & 34.17 & 38.90 & 41.05 & 28.42 & 31.99 & 21.66\\
Mixstyle \cite{zhou2021domain} & 17.61 & 22.53 & 27.54 & 31.64 & 30.42 & 34.11 & 38.71 & 40.86 & 28.57 & 32.29 & 20.35\\
XDED \cite{lee2022cross} & 17.63 & 22.44 & 27.86 & 31.76 & 30.79 & 34.21 & 38.98 & 40.96 & 28.82 & 32.34 & 20.53\\
\hline
CLIPBase & 24.61 & 28.16 & 31.53 & 34.37 & 33.00 & 35.94 & 36.58 & 38.98 & 31.43 & 34.36 & \textbf{11.50}\\
\textbf{SCI-PD} & \textbf{25.28} & \textbf{28.80} & \textbf{33.89} & \textbf{36.30} & \textbf{36.09} & \textbf{38.36} & \textbf{41.36} & \textbf{42.55} & \textbf{34.16} & \textbf{36.50} & 13.64\\
\hline
\end{tabular}
\end{center}
\end{table*}

\subsection{Transferability}
\quad To evaluate the transferability of our method, we conduct experiments using various lightweight vision backbones. We choose EfficientNet-B0 \cite{tan2019efficientnet} and MobileNet-V3 \cite{howard2019searching}, whose parameters are far less than ResNet18 \cite{he2016deep} that are regarded as real-time architectures. Results from Table \ref{tab:trans} present the average accuracy and H-score of four splits and SCI-PD surpasses the baseline method ERM \cite{koltchinskii2011oracle} and CLIPBase with a relatively large margin. Concretely, compared with ResNet18 of 11.4M parameters, EfficientNet-B0 merely has 5.3M parameters but exceeds ResNet18 with 2.43$\%$ and 1.32$\%$ on accuracy and H-score for SCI-PD. Meanwhile, SCI-PD can boost the performance on MobileNet-V3 to achieve comparable results with ERM on ResNet18, but five times less parameters.

\begin{table}[ht]
\small
\setlength{\abovecaptionskip}{-5pt} 
\setlength{\belowcaptionskip}{-3pt} 
\caption{Experimental results of other lightweight vision models on OfficeHome.}
\label{tab:trans}
\setlength{\tabcolsep}{4pt}
\begin{center}
\begin{tabular}{c|c|cc|c}
\hline
\multirow{2}{*}{Method} & \multirow{2}{*}{Params} & \multicolumn{2}{c|}{Average} & \multirow{2}{*}{${\rm H^{2}}$-CV}\\
\cline{3-4}
& & Acc & H-score\\
\hline
EfficientNet-B0 \cite{tan2019efficientnet} & & 58.50 & 53.66 & 9.36\\
+ CLIPBase & 5.3M & 58.71 & 54.23 & 6.28\\
+ \textbf{SCI-PD} & & \textbf{64.16} & \textbf{58.65} & \textbf{5.98}\\
\hline
\hline
MobileNetV3 \cite{howard2019searching} & & 47.47 & 45.00 & 12.13\\
+ CLIPBase & 2.0M & 49.79 & 47.22 & \textbf{8.30} \\
+ \textbf{SCI-PD} & & \textbf{52.74} & \textbf{49.60} & 8.81\\
\hline
\end{tabular}
\end{center}
\end{table}

\subsection{Ablation Study}

\quad \textbf{Key Components.}
As our method is consisted of three types of perturbation, we conduct ablation study to investigate the effectiveness of each component in Table \ref{tab:component}. We start from the vanilla CLIPBase, clarified in Section \ref{sec:method}, as the baseline method. Then we sequentially add instance perturbation, score perturbation and class perturbation. Results show that IP, SP and CP can improve 0.20$\%$, 1.94$\%$ and 1.27$\%$ on H-score. Also, IP and CP improves on ${\rm H^{2}}$-CV that serves as the key to enhance robustness. 

\begin{table}[htbp]
\small
\setlength{\abovecaptionskip}{-5pt} 
\setlength{\belowcaptionskip}{-5pt} 
\caption{Ablation study on different components for SCI-PD on OfficeHome. I-PD denotes perturbation distillation on instance, while SI-PD conducts perturbation on score and instance.} 
\begin{center}
\begin{tabular}{c|ccc|cc|c}
  \cline{0-6}
  \multirow{2}{*}{Model}
  &\multirow{2}{*}{IP} 
  &\multirow{2}{*}{SP} 
  &\multirow{2}{*}{CP}
  &\multicolumn{2}{c|}{Average}
  &\multirow{2}{*}{${\rm H^{2}}$-CV}\\
  \cline{5-6}
  & & & & Acc & H-score & \\
  \hline
  CLIPBase & - & - & - & 58.11 & 53.92 & 6.40 \\
  \hline
  I-PD & \checkmark & - & - & 58.76 & 54.12 & 5.93\\
  SI-PD & \checkmark & \checkmark & - & 60.24 & 56.06 & 6.39\\
  SCI-PD & \checkmark & \checkmark & \checkmark & \textbf{61.73} & \textbf{57.33} & \textbf{5.20}\\
  \hline
\end{tabular}
\label{tab:component}
\end{center}
\end{table}

\begin{figure}[!ht]
  \setlength{\belowcaptionskip}{-5pt}
  \centering
  \subcaptionbox{$\tau$ for SP}{\includegraphics[width=0.32\linewidth]{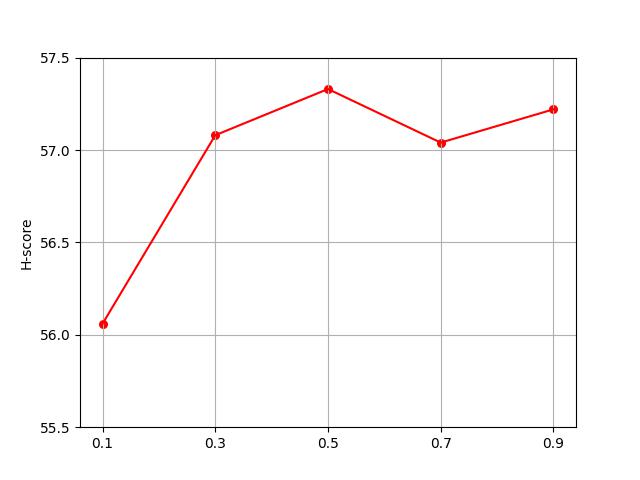}}
  \subcaptionbox{$\alpha$ for IP}{\includegraphics[width=0.33\linewidth]{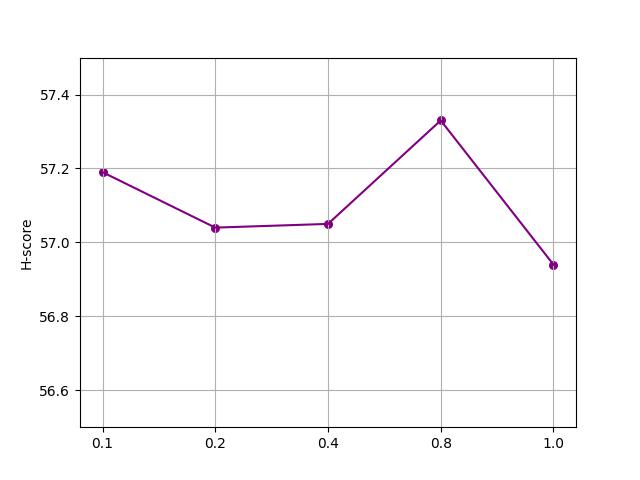}}
  \subcaptionbox{$\beta$ for CP}{\includegraphics[width=0.32\linewidth]{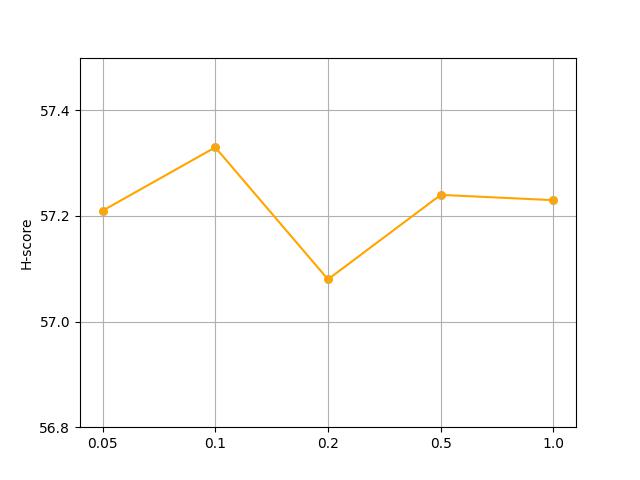}}
  \caption{Experimental results on hyper-parameters.}
  \label{fig:hyper}
\end{figure}

\textbf{Hyper-parameter Analysis.}
Low hyper-parameter sensitivity is a critical determinant for practical applications. Consequently, we conduct experiments on $\tau$, $\alpha$ and $\beta$ to demonstrate the practicality of our method. Fig. \ref{fig:hyper} shows the average H-score on four splits. The fluctuation of three hyper-parameters are low with merely 1.27$\%$, 0.39$\%$ and 0.25$\%$. The best performance is achieved when $\alpha=0.8$, $\tau=0.5$ and $\beta=0.1$.

\textbf{Other Variants.}
Most recent studies on VLMs for the DG task adopt the fine-tuning paradigm that the downstream model should have identical architecture as VLMs. Nevertheless, we extend SCI-PD to larger vision models for comparison with SOTA methods on \cite{cha2022domain} and the zero-shot ability of CLIP models \cite{radford2021learning}. For a fair comparison, we all select CLIP model with ResNet50 as the image encoder. Results in Table \ref{tab:variant} show that our method achieves a stable performance that the ${\rm H^{2}}$-CV is merely 3.04$\%$, exceeding MIRO \cite{cha2022domain} of 11.76$\%$. Especially when $\mathcal{H}=0$, SCI-PD surpasses MIRO with 20.05$\%$ on accuracy and 15.03$\%$ on H-score.


\begin{table}[ht]
\small
\setlength{\abovecaptionskip}{-5pt} 
\setlength{\belowcaptionskip}{-3pt} 
\caption{Comparison of methods on zero-shot and re-training paradigm for OfficeHome.}
\label{tab:variant}
\begin{center}
\begin{tabular}{c|c|cc|c}
\hline
\multirow{2}{*}{Method} & \multirow{2}{*}{Type} & \multicolumn{2}{c|}{Average} & \multirow{2}{*}{${\rm H^{2}}$-CV}\\
\cline{3-4}
& & Acc & H-score\\
\hline
CLIP \cite{radford2021learning} & Zero-shot & 51.31 & 50.53 & - \\
MIRO \cite{cha2022domain} & Re-train & 56.56 & 51.90 & 14.80\\
\textbf{SCI-PD} & Distill & \textbf{63.48} & \textbf{57.30} & \textbf{3.04}\\
\hline
\end{tabular}
\end{center}
\end{table}

\subsection{Visualization}
\begin{figure}[!ht]
  \setlength{\belowcaptionskip}{-5pt}
  \centering
  \subcaptionbox{ERM \cite{koltchinskii2011oracle}}{\includegraphics[width=0.33\linewidth]{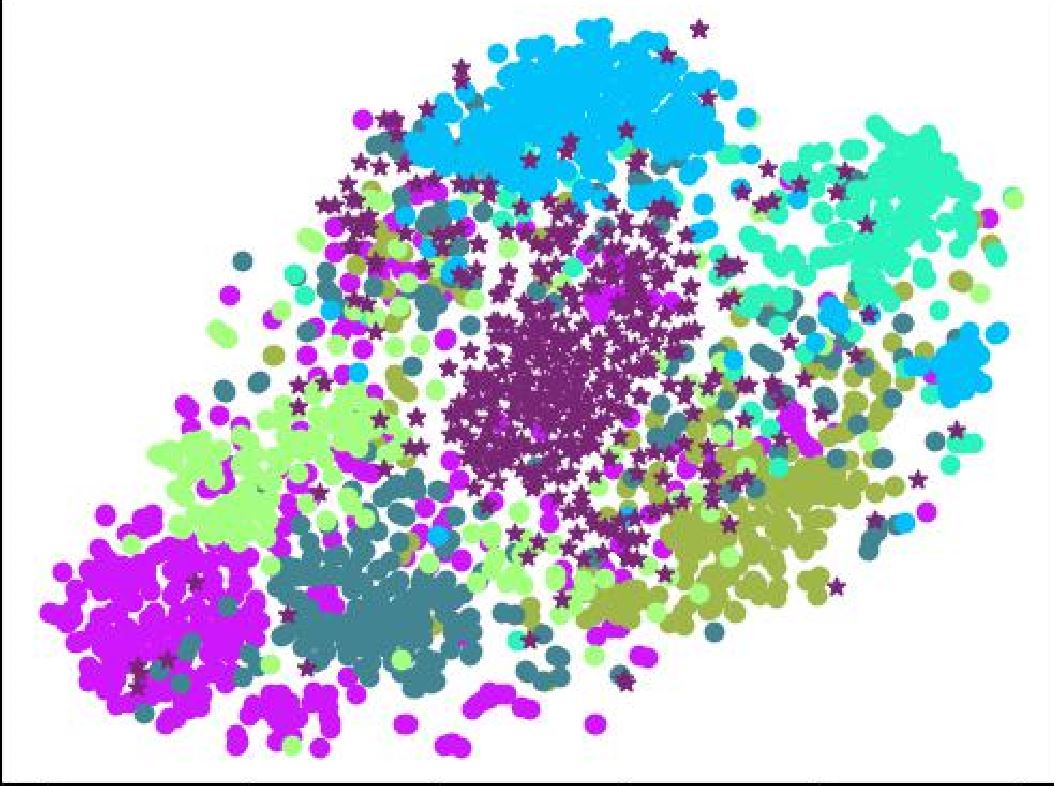}}
  \subcaptionbox{CORAL \cite{sun2016deep}}{\includegraphics[width=0.33\linewidth]{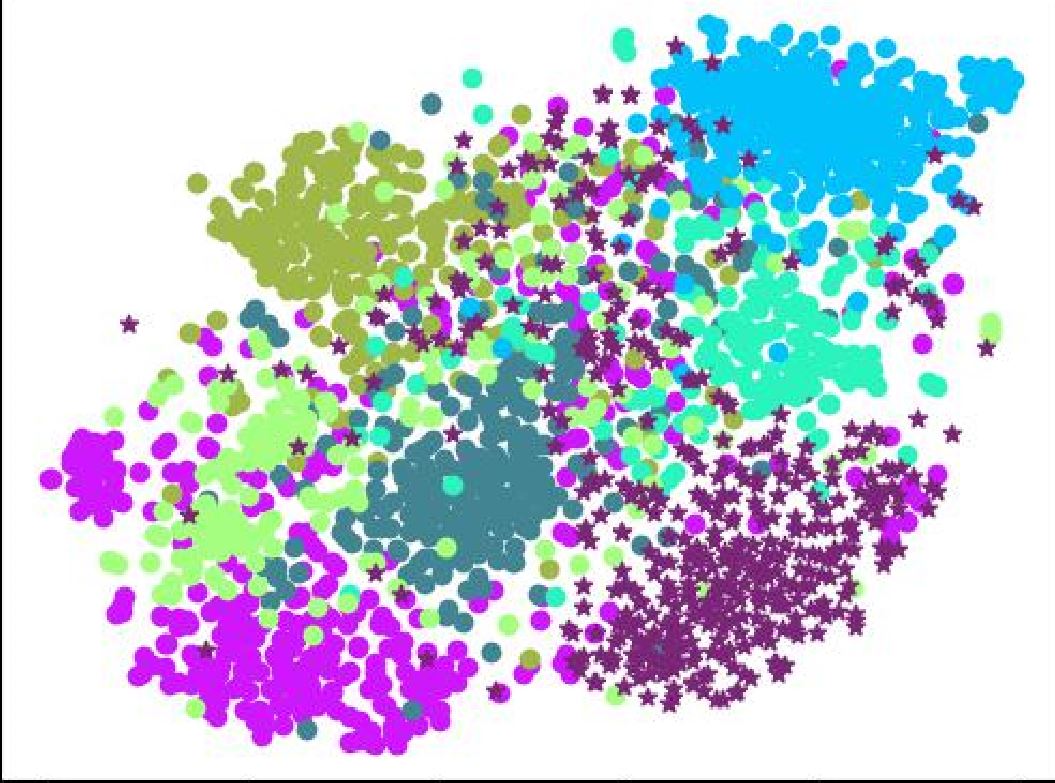}}
  \subcaptionbox{SCI-PD}{\includegraphics[width=0.32\linewidth]{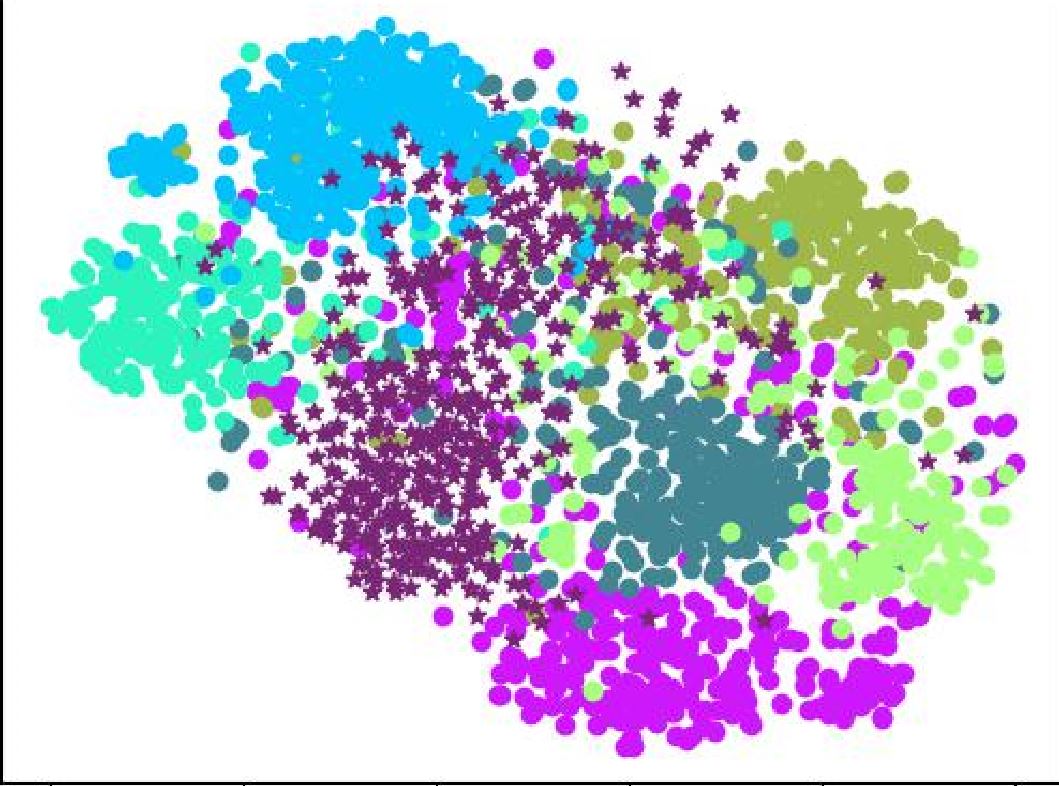}}
  \caption{The t-SNE \cite{van2008visualizing} results of feature distribution on PACS when $\mathcal{H}=0$.}
  \label{fig:vis}
\end{figure}

\quad We present the t-SNE \cite{van2008visualizing} visualization of the feature distribution on PACS when $\mathcal{H}=0$. As displayed in Fig. \ref{fig:vis}, for the baseline method ERM \cite{koltchinskii2011oracle}, the boundary between categories is ambiguous. The efficient method CORAL \cite{sun2016deep} has improved the compactness between clusters, but the performance on unknown categories declines. Nevertheless, with the guidance on CLIP model \cite{radford2021learning}, SCI-PD can promote the intra-class compactness and inter-class variance that improves the performance.

\vspace{-0.3cm}
\section{Conclusion}
\quad In this paper, we investigate the issues in practical scenarios of domain generalization. We firstly develop a novel Perturbation Distillation (PD) algorithm, to transfer zero-shot ability from vision-language models to lightweight vision models, thereby avoiding large computation costs in conventional fine-tuning paradigm. We introduce the perturbation from Score, Class and Instance (SCI) that sufficiently excavate the knowledge from VLMs. Furthermore, we propose a Hybrid Domain Generalization (HDG) benchmark and a novel metric ${\rm H^{2}}$-CV to comprehensively evaluate the model robustness. Experimental results demonstrate that our method achieves the state-of-the-art performance with a relatively large margin on three diverse metrics.

\section{Acknowledgment}
\quad This work is supported by Chinese National Natural Science Foundation under Grants (62076033).
{
    \small
    \bibliographystyle{ieeenat_fullname}
    \bibliography{main}
}


\end{document}